\documentclass[letterpaper,10pt,journal,table]{ieeeconf}

\usepackage{graphics} 
\usepackage{todonotes}
\usepackage{subfiles}
\usepackage{verbatim}
\usepackage{colortbl}
\usepackage{multicol}
\usepackage{graphicx}
\usepackage{cuted}
\usepackage{multirow}
\usepackage{caption}
\usepackage{floatrow}
\usepackage{subcaption}
\usepackage{hyperref}
\usepackage{tabularx,booktabs}

\usepackage{siunitx}
\DeclareSIUnit{\deg}{deg}
\DeclareSIUnit{\pixel}{px}
\DeclareSIUnit{\fps}{fps}

\newcommand{\eg}{{\emph{e.g.}}}
\newcommand{\ie}{{\emph{i.e.}}}

\newcommand{\etc}{{\emph{etc.}}}

\newcommand{\thedataset}{PedSynth}
\newcommand{\thefw}{ARCANE}

\title{Synthetic Data Generation Framework, Dataset, and Efficient Deep Model for Pedestrian Intention Prediction}

\author{Muhammad Naveed Riaz$^{1}$, Maciej Wielgosz$^{2}$, Abel García Romera$^{1}$, and Antonio M. López$^{1}$ 
\thanks{Corresponding author: {\tt\small nriaz@cvc.uab.cat}} %
\thanks{$(1)$ Naveed, Abel, and Antonio are with the Dpt. Ciències de la Computació and the CVC, at Univ. Autònoma de Barcelona (UAB).} %
\thanks{Naveed acknowledges the support for his PhD research provided by the FI fellowship AGAUR 2021 FI-SDUR 00281 (Secretaria d’Universitats i Recerca of the Generalitat de Catalunya and the Fons Social Europeu). Naveed and Antonio acknowledge the support provided by SGR 2021 ADAS (ref. 2021 SGR 01621; Departament de Recerca i Universitats of the Generalitat de Catalunya) during the FI fellowship application.}
\thanks{Maciej acknowledges the funding from the European Union’s Horizon 2020 research and innovation programme under Marie Skłodowska-Curie grant agreement No. 801342 (Tecniospring INDUSTRY) and the Government of Catalonia’s Agency for Business Competitiveness (ACCIÓ). This allowed to develop ARCANE. This work only expresses the opinion of the author and neither the European Union nor ACCIÓ are liable for the use made of the information provided.} %
}

\begin{document}

\onecolumn

\maketitle
\thispagestyle{empty}
\pagestyle{empty}

\begin{abstract}
Pedestrian intention prediction is crucial for autonomous driving. In particular, knowing if pedestrians are going to cross in front of the ego-vehicle is core to performing safe and comfortable maneuvers. Creating accurate and fast models that predict such intentions from sequential images is challenging. A factor contributing to this is the lack of datasets with diverse crossing and non-crossing (C/NC) scenarios. We address this scarceness by introducing a framework, named {\thefw}, which allows programmatically generating synthetic datasets consisting of C/NC video clip samples. As an example, we use {\thefw} to generate a large and diverse dataset named {\thedataset}. We will show how {\thedataset} complements widely used real-world datasets such as JAAD and PIE, so enabling more accurate models for C/NC prediction. Considering the onboard deployment of C/NC prediction models, we also propose a deep model named PedGNN, which is fast and has a very low memory footprint. PedGNN is based on a GNN-GRU architecture that takes a sequence of pedestrian skeletons as input to predict crossing intentions. {\thefw}, {\thedataset}, and PedGNN will be publicly released\footnote{\scriptsize{$<$URL address to be provided with the camera-ready version of the paper$>$}}.
\end{abstract}



\section{Introduction}
As evidenced in an early Google self-driving car report \cite{google2015disengagements}, the 10\% of their self-driving malfunctions on streets were due to incorrect behavior predictions of other road users, including pedestrians.
\begin{figure}[thpb]
      \centering 
      \includegraphics[width=1.0\columnwidth]{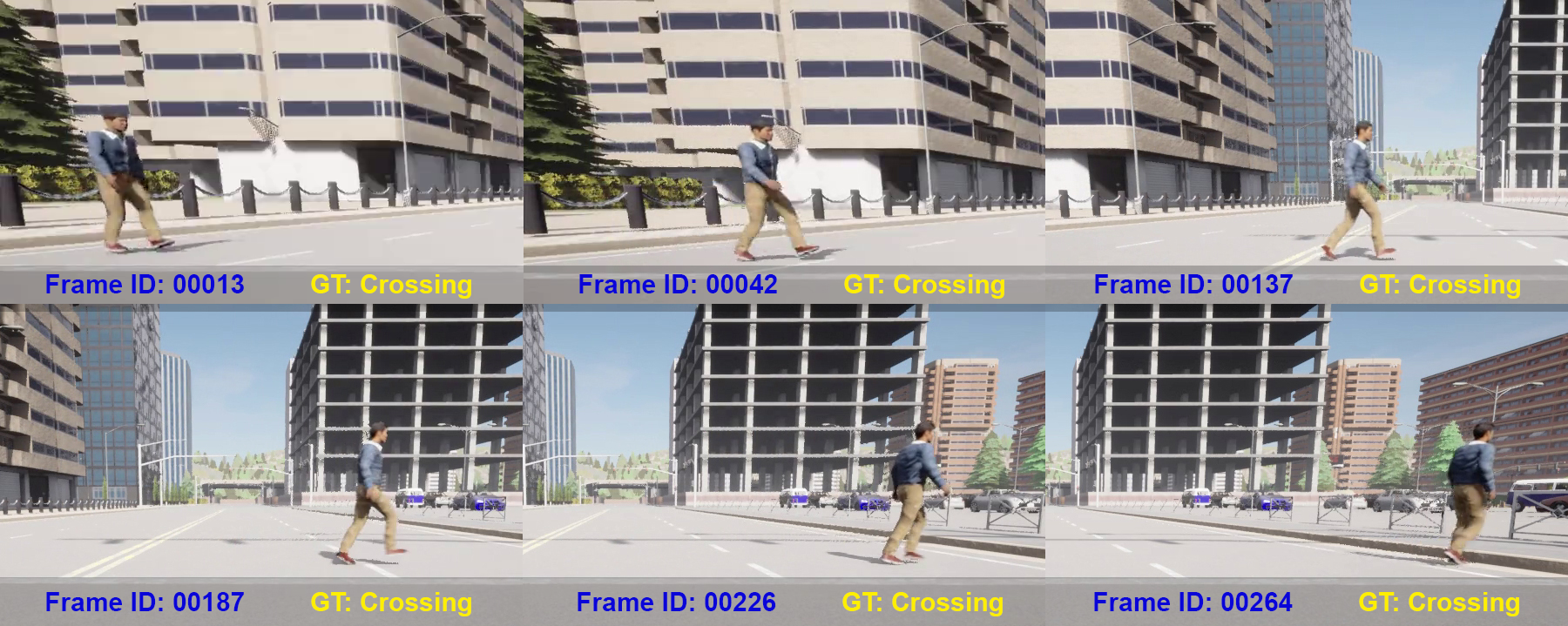}
      
      \vspace{0.1cm}
      
      \includegraphics[width=1.0\columnwidth]{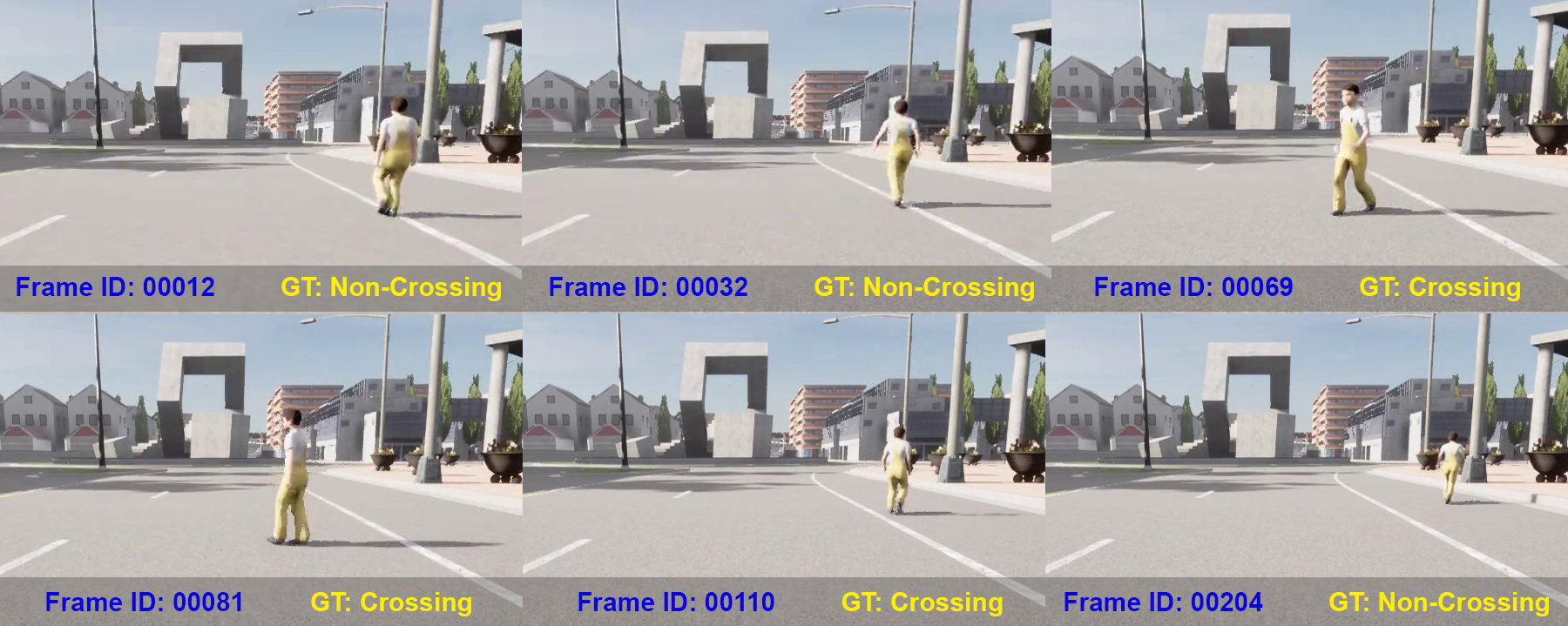}
      \captionof{figure}{Summary of two video clips from {\thedataset}. Top rows: a pedestrian crosses the road perpendicularly to the ego-vehicle moving direction. Bottom rows: a pedestrian change the intention of crossing the road at mid-lane. In both examples, the pedestrians enter the road at locations not enabled for crossing.}
      \label{fig:visualabstract}
\end{figure}
While there have been significant efforts to improve the accuracy of pedestrian intention prediction \cite{fang2018IsTP, bouhsain2020pedestrian, gesnouin2021TrouSPINetSA, kotseruba2021benchmark, ham2022MCIP, wang2022pedestrian, cadena2022pedgraph+}, there is still ample room for improvement. Currently, two datasets, JAAD \cite{rasouli2017jaad} and PIE \cite{rasouli2019PIE}, are being used to benchmark such prediction models. In these datasets, the core ground truth (GT) consists of labeling if pedestrians are crossing or are going to cross in front of the ego vehicle. As for other onboard perception tasks ({\eg}, object detection and tracking \cite{cabon2020VKITTI2}, semantic segmentation \cite{wrenninge2018synscapes}, monocular depth estimation \cite{gurram2021monoDEVS}), synthetic datasets have been proposed to train C/NC prediction models \cite{achaji2022isattention, bai2022DeepVD}. We propose to go beyond these datasets by introducing a framework, named {\thefw}\footnote{{\thefw} stands for \emph{\textbf{a}dve\textbf{r}sarial} \emph{\textbf{c}ases} for \emph{\textbf{a}uto\textbf{n}omous} \emph{v\textbf{e}hicles}, the generic project supporting the development of the framework.}, where traffic scenarios of pedestrian behavior can be programmatically defined. This opens the possibility of introducing underrepresented vehicle-to-pedestrian traffic situations. For being aligned with the research community, {\thefw} has been developed on top of the CARLA simulator \cite{dosovitskiy17CARLA}. As an example, we have used {\thefw} to generate {\thedataset} which is a large and diverse synthetic dataset with pedestrian C/NC labels. Note that this type of labeling is not provided by the CARLA simulator, but it is generated by {\thefw}. {\thedataset} consists of 947 video clips of pedestrian C/NC situations. Each video clip runs $\sim20$s at 30fps, so resulting in approximately 5 H and 26 min of labeled videos. Figure \ref{fig:visualabstract} shows several frames of two video clips from {\thedataset}. On the other hand, users can generate their own datasets by working with {\thefw}. 

Focusing on the demanding hardware requirements for onboard perception, we also propose a lightweight model for C/NC prediction, named PedGNN. This model has a 27KB GPU memory footprint and runs on $\sim0.6$ms on an NVIDIA GTX 1080 GPU. Compared to a state-of-the-art C/NC prediction model, here named PedGraph+ \cite{cadena2022pedgraph+}, PedGNN is one order of magnitude smaller and one order of magnitude faster. Even though, PedGNN outperforms PedGraph+ in terms of the class-balanced F1-score classification metric. PedGNN is based on a GNN-GRU architecture that takes a sequence of pedestrian skeletons as input to predict crossing intentions (see Fig. \ref{fig:model}). Note that, so far, the spatiotemporal analysis of pedestrian skeletons has been shown as one of the most relevant sources of information to predict pedestrian crossing intentions \cite{fang2018IsTP, gesnouin2021TrouSPINetSA, bai2022DeepVD, cadena2022pedgraph+}. 

Using PedGNN and {\thedataset} to complement the training sets of both JAAD and PIE, allows us to outperform pedestrian C/NC prediction in the respective testing sets. 

\section{Related Work}
\label{sec:rw}

\subsection{Pedestrian intention prediction}
Pioneering approaches cast C/NC prediction as a trajectory prediction problem, which requires the explicit estimation of the future location, speed, and acceleration of the observed pedestrians \cite{schneider2013pedestrian, keller2014will}. In practice, the corresponding dynamic models were difficult to adjust and require to extract the silhouette of the pedestrians, dense depth, and dense optical flow with ego-motion compensation. On the other hand, Schneemann and Heinemann \cite{schneemann2016context} concluded that pedestrians’ posture and body movement are essential to take faster C/NC predictions. Accordingly, methods relying on the temporal evolution of pedestrian skeletons gained popularity, especially due to the increasing accuracy of the deep models adjusting them in 2D images, {\eg}, see \cite{cao2017realtime} (becoming OpenPose \cite{cao2021openpose}) and \cite{fang2017rmpe} (becoming AlphaPose \cite{fang2023alphapose}). For instance, a sequence of pedestrian skeletons was used as input to classical lightweight and fast machine learning models such as a random forest C/NC classifier \cite{fang2018IsTP} and as input to more resources-consuming but accurate deep models such as a Graph Convolutional Network (GNN) \cite{cadena2019pedgraph}. Skeleton extraction and C/NC prediction have been also tackled as a joint multi-task problem \cite{razali2021pedestrian}.

Semantic and contextual information is also considered in different works. In \cite{kotseruba2021benchmark}, it is proposed a deep model based on recurrent neural networks (RNNs) and attention modules, which takes as input the ego-vehicle speed, the bounding box (BB), skeleton, and local context (RGB crops) of pedestrians. In \cite{yang2021predicting}, RNNs and attention modules are also used to train two intermediate deep architectures whose output is fused (mid/late fusion) to provide C/NC predictions. One architecture considers the ego-vehicle speed, the BB, and the skeleton of pedestrians. The other considers local and global (scene semantic segmentation) contexts. Finally, \cite{cadena2019pedgraph} became a state-of-the-art model \cite{cadena2022pedgraph+}, here named PedGraph+, by incorporating the ego-vehicle speed and pedestrian local context to the initial skeleton-based GNN architecture. 

\emph{In this paper:} As these C/NC prediction approaches, we also rely on an off-the-self model to obtain the pedestrian skeletons that PedGNN requires as input. Since these skeletons are structured as graphs, we think that GNNs are natural architectures to work with them, {\ie}, as done by PedGraph+. However, we want PedGNN to be more lightweight and faster. Thus, we use a different GNN architecture than PedGraph+. 

\subsection{Synthetic datasets focusing on C/NC prediction}

As with other vision-based tasks, C/NC prediction research started with relatively small and non-naturalistic datasets \cite{schneider2013pedestrian, fang2018IsTP}. Fortunately, larger and naturalistic datasets such as JAAD \cite{rasouli2017jaad} and PIE \cite{rasouli2019PIE} appeared progressively, so helping to accelerate this research. It was also a matter of time to use synthetic data to support C/NC prediction research. In fact, onboard pedestrian detection was one of the first tasks for which a model was trained on synthetic images to perform later in real-world images, this was done more than a decade ago \cite{marin2010learning}. Since then, there have been many works leveraging synthetic data to support the training of perception models or performing simulations \cite{ros2016SYNTHIA, richter2016playing, wrenninge2018synscapes, cabon2020VKITTI2, dosovitskiy17CARLA, shah2018airsim}; being synth-to-real domain adaptation a core ingredient to encourage the use of synthetic data \cite{wang2018survey,csurka2022semsegreview}. 

Focusing on pedestrians, synthetic data has been mainly generated and used for the tasks of detection and tracking either onboard or from static infrastructure locations \cite{hattori2015learning, bochinski2016training, cheung2017MixedPedsPD, kim2018PedXBD, stauner2022SynPeDS}; where the required GT for each pedestrian consists of a 2D/3D BB, pixel-level segmentation and depth, an ID, and, eventually, a body skeleton. In addition to this kind of GT, for collecting samples to develop C/NC prediction models we must control pedestrian behavior in the simulator, {\eg}, to force C/NC situations as we wish, and we must label each frame accordingly as in Figure \ref{fig:visualabstract}. Recent attempts to do so \cite{achaji2022isattention, bai2022DeepVD} rely on the CARLA simulator \cite{dosovitskiy17CARLA}. In \cite{achaji2022isattention}, the CP2A dataset was introduced with 220K video clips with per-frame C/NC labels, where 25\% of the clips contain crossing (c) examples. In \cite{bai2022DeepVD}, the Virtual-Pedcross-4667 dataset was introduced with 4,667 video clips specially prepared to cover a variety of weather and lighting conditions and per-frame C/NC labels, where 61\% of the clips contain crossing (c) examples. On the other hand, these datasets lack some corner cases like the one shown as the bottom example in Figure \ref{fig:visualabstract}. Overall, the experiments provided in  \cite{achaji2022isattention, bai2022DeepVD} encourage the use of synthetic data to train C/NC models. 

\emph{In this paper:} We contribute to the use of synthetic data to develop C/NC prediction models. As \cite{achaji2022isattention, bai2022DeepVD} we rely on the CARLA simulator. However, rather than only providing a specific synthetic dataset, we introduce {\thefw}, a framework prepared to programmatically generate synthetic datasets of pedestrian C/NC videos. As an example, we have used {\thefw} to generate {\thedataset}, which consists of 
947 video clips recorded under different weather and lighting conditions over 400 locations in CARLA cities, with $\sim398$K frames with C/NC labels. Moreover, using PedGNN, we show that {\thedataset} is a good complement for the training sets of both JAAD and PIE, so boosting C/NC prediction performance in the respective testing sets.

\begin{figure*}[t!]
\centering 
\includegraphics[width=\textwidth]{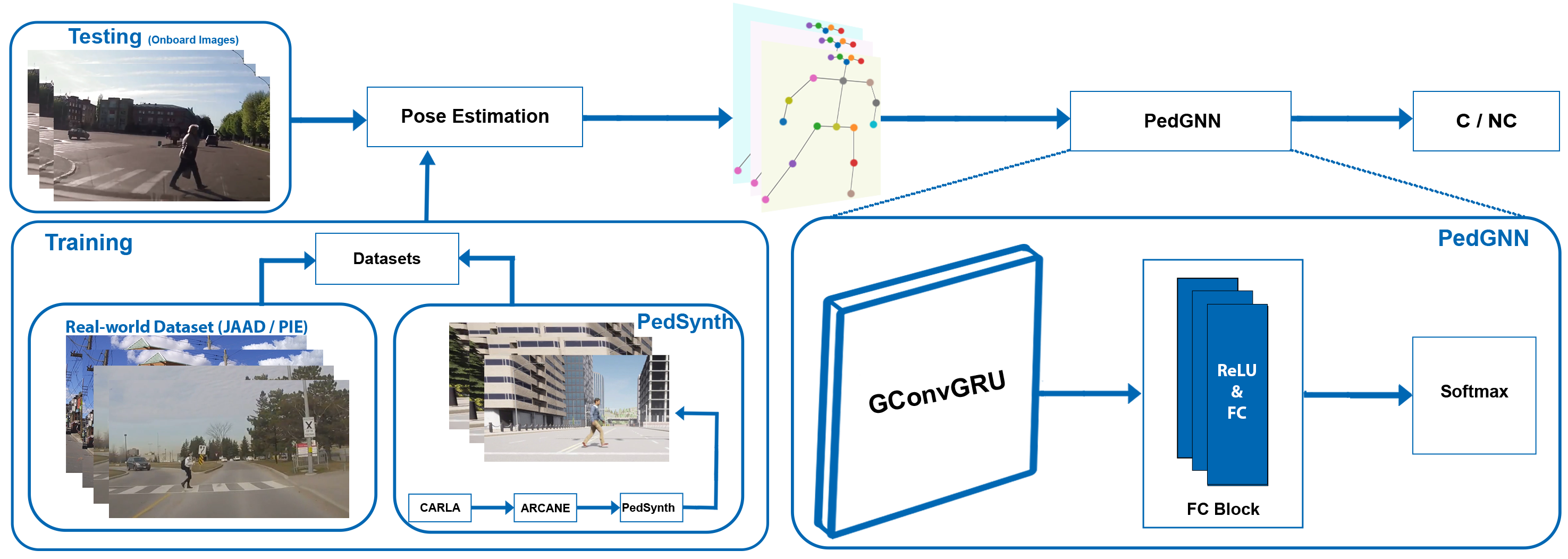} 
\captionof{figure}{To perform C/NC predictions PedGNN processes sequences of pedestrian skeletons. To process onboard sequences while driving, we use a temporal sliding window of a 1-frame step. PedGNN consists of a graph convolutional gated recurrent unit (GConvGRU), followed by a block of three (ReLU + Fully connected) layers, and a final Softmax. Synthetic datasets with C/NC examples can be used for training PedGNN. For instance, in this paper, we use {\thedataset}, a synthetic dataset that we have generated using {\thefw}, a framework that we introduce in this paper too (see Fig. \ref{fig:arcane}).} 
\label{fig:model} 
\end{figure*}

\begin{figure}[t!]
\centering 
\includegraphics[width=\textwidth]{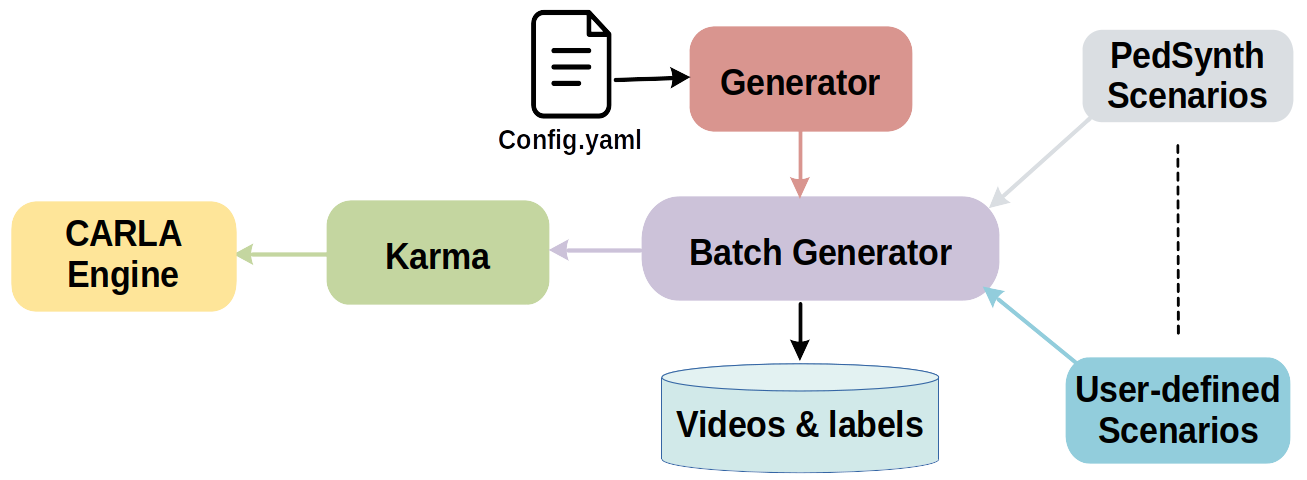} 
\captionof{figure}{Block diagram of ARCANE dataset generator. } 
\label{fig:arcane} 
\end{figure}

\section{Methods}
\label{sec:methods}

Figure \ref{fig:model} summarizes the role of the main contributions of this paper: {\thefw}, {\thedataset}, and PedGNN, which we present in the following subsections.

\subsection{The {\thefw} framework}
{\thefw} framework is built on the top of CARLA simulator. It enables the generation of different types of video clips through the parameterization of the distribution of pedestrian models, pedestrian velocity, and onboard camera settings. It is possible to create scenarios in a single Python file, which can establish the trajectories of pedestrians in the scene. {\thefw} also contains a series of mechanisms that allow for the filtering of not useful videos ({\eg}, when a pedestrian is not visible). {\thefw} is structured around five primary modules, as shown in Fig. \ref{fig:arcane}.


    \textbf{Generator}: This module manages the simultaneous execution of multiple batch generator objects. It oversees the randomization of the data generation process and finalizes the process once completed. The module maintains a count of the number of generated videos, ensuring the target quantity is achieved. In the event of crashes, the module has a predefined number of retry attempts to prevent endless generations.

    \textbf{Batch Generator}: This module spawns pedestrians in a city, positions cameras, and regulates the data generation process. An integral part of its role involves verifying the presence of pedestrians in the generated videos. This is done via semantic segmentation checks and skeleton existence verification. It supervises the C/NC labeling process too, {\eg}, marking pedestrians as crossing (C) if they are entering a driving area.
    These tasks are fulfilled by interacting with Karma.
   
    \textbf{Karma}: 
    This module acts as a facade for the CARLA API. It automates the process of creating the virtual environment in CARLA, pedestrian spawning, and other tasks, by relying on CARLA-based scenario runner functionalities.
    
    \textbf{CARLA Engine}: This module serves as an instance of the CARLA simulator, which is supplemented with extra routines to ensure its operation within a Docker container during the data generation phase. The module is equipped to restart the container in case of any simulation crashes.



Based on these modules, {\thefw} allows for both \emph{simple} and \emph{advanced usage}. Advanced usage refers to the possibility of programmatically defining dynamic traffic scenarios. For instance, leveraging CARLA cities it is possible to write a Python code to choreograph the behavior of pedestrians in these towns, so forcing situations interesting for C/NC prediction. This is what we have done for generating the {\thedataset} dataset, as illustrated in Fig. \ref{fig:arcane}. Given one of such user-defined Python files to generate traffic scenarios, it is possible to generate variations by generic parameters ({\ie}, scenario agnostic) which allow controlling the types of pedestrians to be included, their speed, the pedestrian density, {\etc} These parameters are included in a configuration file, named {\tt config.yaml} in Fig. \ref{fig:arcane}. Therefore, this configuration file enables simple usage provided we are satisfied with the traffic scenarios in place. 






Overall, the development of {\thefw} took approximately half a year. The code repository includes 3,267 lines of Python code in 47 files, supplemented by a multitude of additional files of other formats.

\newcommand{\localtimes}{\!\!\times\!\!}
\begin{table}[t!]
\caption{Features of the datasets used in this paper.}
\label{tab:statsdataset}
\begin{center}
\small
\setlength{\tabcolsep}{1pt}
\begin{tabular}{@{}lccc@{}}
\toprule
Feature                         & JAAD       & PIE        & {\thedataset} \\
\midrule
Video clips with C/NC labels         & 323        & 55         & 947   \\
Video clips length (s)          & $\sim5-15$ & $\sim600$  & $\sim20$     \\
Frames per second (fps)         &  30        &  30        & 30     \\
Frame resolution (pix)          &  $1920\localtimes1080$   &  $1920\localtimes1080$   &  $1600\localtimes600$    \\
Frames with C/NC labels        & $\sim75$K  & $\sim293$K & $\sim398$K \\
Semantic segmentation           & no         & no          & yes  \\
Pedestrian skeleton             & no         & no          & yes  \\
Weather variability             & yes        & no          & yes  \\
\bottomrule
\end{tabular}
\end{center}
\end{table}
\let\localtimes\undefined

\subsection{The {\thedataset} dataset}
We have written a Python code, named {\tt {\thedataset} Scenarios} in Fig. \ref{fig:arcane}, which is consumed in {\thefw} to generate video clips with C/NC labels according to the settings provided through the {\tt config.yaml} file. With this information, {\thefw} has generated the {\thedataset} dataset. It covers $\sim400$ locations from different CARLA cities, thus, including different city styles and road lanes. Varying pedestrians and environmental conditions, we have generated 947 video clips with C/NC labels, resulting in a total of $\sim398$K frames with C/NC labels. Table \ref{tab:statsdataset} summarizes the main features of {\thedataset} compared to the real-world datasets JAAD and PIE.  We can see how {\thedataset} contains $\sim100$K more frames with C/NC labels than PIE and more than $\sim300$K compared to JAAD.  Note that, as in real-world datasets, {\thedataset}'s videos include frames with no pedestrians, where C/NC prediction models should not rise false warnings. Beyond C/NC labels, we can also leverage GT already present in the CARLA simulator itself, such as pixel-level class semantics (semantic segmentation), pedestrian skeletons, {\etc} We provide more detailed information about {\thefw} and {\thedataset} in the technical report \cite{wielgosz2023carla}.

\begin{figure}
  \centering
  \includegraphics[width=0.3\columnwidth ]{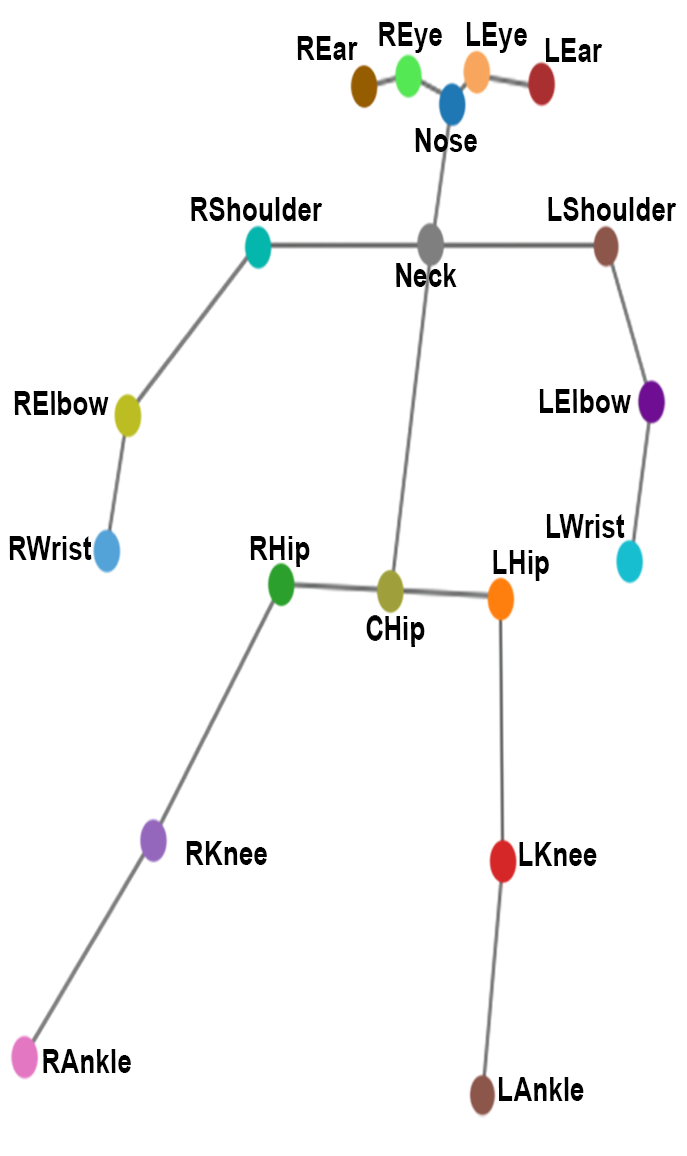}
  \caption{Pedestrian skeleton as expected by PedGNN. We consider 19 joints connected as an undirected graph.}
  \label{fig:skeletonjoints}
\end{figure}

\subsection{PedGNN model}
As we have mentioned in Section \ref{sec:rw}, the temporal evolution of pedestrian pose is considered core information to determine C/NC intentions. Today, there are robust deep models able to provide human-body skeletons from 2D images \cite{cao2021openpose,fang2023alphapose}. Even by using hand-crafted features and traditional machine learning models, the temporal evolution of 2D-fitted pedestrian skeletons was shown to be effective to determine C/NC intentions \cite{fang2018IsTP}. Therefore, as per the state-of-the-art literature, for our C/NC prediction model, we also assume that a sequence of 2D-fitted pedestrian skeletons is used as input to determine C/NC intentions. Figure \ref{fig:skeletonjoints} shows the joints we consider and their connections. They cover the head, arms, trunk, and legs. The poses of hands and feet are not considered since they cannot be clearly perceived by an onboard camera, and most likely they are irrelevant for determining C/NC intentions. To process sequences of images in a continuous manner, we use a temporal sliding window approach with a frame step to be adjusted experimentally according to the frame rate of the onboard camera ({\eg}, we use a 1-frame step for cameras working at 30fps).

Since pedestrian skeletons can be represented as undirected graphs, natural deep architectures to process them are GNNs (graph neural networks). In fact, since, for each pedestrian, we work with a sequence of skeletons, a graph convolutional gated recurrent unit (GConvGRU) is a very convenient model for C/NC prediction. Therefore, we adopt it by using the implementation in the PyTorch Geometric (PyG) library\footnote{\url{https://github.com/pyg-team/pytorch_geometric}}. The output of the GConvGRU is flattened and processed by three consecutive blocks of (ReLU + FC) layers, which feed Softmax to obtain the C/NC prediction (see Fig. \ref{fig:model}). 

As input information at a graph node, we use the $(x_j,y_j)$ coordinates of the joint $j$ associated with the node and its fitting confidence $c_j$ as provided by the skeleton fitting model in use. As is recommended for GNNs \cite{kipf2017semisupervised} and for skeleton-based C/NC prediction \cite{fang2018IsTP}, $(x_j,y_j)$ are normalized at each frame to the range $[0,1]$. Thus, we work with normalized 2D coordinates $(\hat{x}_j,\hat{y}_j)$ which add invariance to ego-vehicle to pedestrian distance variations. Overall, the input to PedGNN has dimensions ($N_F$,19,3), where $N_F$ is the number of frames used to perform C/NC predictions, which is determined experimentally during the training of PedGNN. Obviously, the 19 comes from the number of joints, and 3 from the information per joint, {\ie}, $(\hat{x}_j,\hat{y}_j,c_j)$.

Finally, we remark that we focus on having a lightweight and fast C/NC prediction model. PedGNN shows a memory footprint of 27KB and an inference time of $\sim0.6$ms on an NVIDIA GTX 1080 GPU. As we will see in Section \ref{sec:experiments}, this is one order of magnitude of improvement over other state-of-the-art methods such as \cite{cadena2022pedgraph+}.

\section{Experimental Results}
\label{sec:experiments}


\begin{table}[t!]
\caption{Let's a \emph{sample} be a particular pedestrian completing a C/NC sequence. Thus, different samples can overlap in the same frame. Let $N_F^S$ be the number of labeled frames of the C/NC sequence of sample $S$. Let $N_S$ be the number of samples in a particular subset of videos. For each dataset and split subset, this table reports $\#l=\sum_{s=1}^{N_S}\sum_{f=1}^{N_F^s}gt(s,f,l)$, where $l\in\{\mbox{C, NC}\}$ is the label, and $gt(s,f,l)=1$ if $l$ matches the C/NC GT associated to the pedestrian sample $s$ at frame $f$, and $gt(s,f,l)=0$ otherwise.}
\label{tab:labelsdataset}
\begin{center}
\small
\setlength{\tabcolsep}{1pt}
\begin{tabular}{@{}l|rrr|rrr|rrr@{}}
\toprule
Int.  & \multicolumn{3}{c|}{JAAD} & \multicolumn{3}{c|}{PIE} & \multicolumn{3}{c}{\thedataset} \\
Label & Train  & Val. & Test      & Train   & Val.  & Test   & Train   & Val.   & Test \\
\midrule
\#C  & 39.7K & 6.3K  & 32.8K    &  116.2K & 18.1K & 130.9K &  155.1K & 50.4K & 50.5K \\
\#NC &  7.9K & 1.5K  &  8.4K    &  110.7K & 22.1K &  76.8K &   82.3K & 29.9K & 29.5K \\
\bottomrule
\end{tabular}
\end{center}
\end{table}

\subsection{Datasets, metrics, frameworks, pose estimation}
For our experiments, as real-world datasets we use the de-facto standards for C/NC prediction, {\ie}, JAAD \cite{rasouli2017jaad} and PIE \cite{rasouli2019PIE}. We use JAAD \emph{all} version. As a synthetic dataset, we use\footnote{At the moment of elaborating our work, CP2A and Virtual-Pedcross-4667 datasets do not seem downloadable anymore.} our {\thedataset}. Table \ref{tab:statsdataset} summarizes their main features. 
For JAAD and PIE datasets we also use their standard Train/Val./Test split. For {\thedataset} we performed a random split and fix it for all the experiments. Specifically, $\sim80$\% of {\thedataset} is allocated for training, $\sim10$\% for validation, and another $\sim10$\% for testing. Table \ref{tab:labelsdataset} provides information on the corresponding splits in terms of C/NC labeling.

To report our results, we apply the metrics used in C/NC prediction literature, {\ie}, standard \emph{Accuracy}, \emph{Precision}, \emph{Recall}, and \emph{F1-score}. For training models and running inferences, we use PyTorch. 
Since PedGNN is based on pedestrian skeletons adjusted on 2D images, we use the state-of-the-art deep model named AlphaPose \cite{fang2023alphapose} as an off-the-shelf method. AlphaPose does not return the 19 joints we use for PedGNN. Compared to Fig. \ref{fig:skeletonjoints}, AlphaPose does not provide the Neck and CHip joints. To compute the Neck coordinates we average LShoulder and RShoulder coordinates. Analogously, to compute the CHip coordinates we average LHip and RHip.

\subsection{Training protocol}
As the training optimizer, we use AdamW with binary cross-entropy loss and default parameters except for the learning rate, $l_r$. We perform training runs for a maximum number of $M_E$ epochs. We also apply a 50\% dropout. In this optimization process, a training sample consists of a sequence of skeletons from the same pedestrian. In other words, for C/NC pedestrian prediction, we consider $N_F$ consecutive frames. We use a training batch size of 500 samples. Since skeleton information has a really low memory footprint, this batch size fits well in a single 24GB memory GPU. Since some experiments rely on training images from different datasets,  we utilized PyTorch's ConcatDataset and WeightedRandomSampler functions to ensure equal dataset sampling per batch.


To train a C/NC prediction model, we test different values for $N_F$ and $l_r$. Regarding $N_F$, we consider values in the range $[4,\ldots,32]$ with step=2. Since all datasets were recorded at 30fps (Table \ref{tab:statsdataset}), this is equivalent to considering a temporal window from $\sim133$ms to $\sim1067$ms. Regarding $l_r$, we consider values in $\{0.001, 0.005, 0.0002, 0.0005\}$. While training a model, we assess its F1-score at the end of each epoch with the help of the validation set associated with the targeted training set. We have set $M_E=100$. To apply the C/NC prediction models we use a temporal sliding window with a step of 1 frame. Across the considered $N_F$ and $l_r$ values and epochs, we take as the final model the best-performing one. 

Depending on the size of the training set, obtaining a trained model requires from $\sim3-4$h (single training dataset) to $\sim6-9$h (multiple training datasets) in a desktop PC with an NVIDIA RTX 3090 GPU. We remark that, for doing these experiments, pedestrian skeletons are computed and recorded on the hard disk once. 

\newcommand{\vb}[1]{\textbf{#1}} 
\newcommand{\vs}[1]{\textit{#1}} 
\begin{table}
\caption{C/NC prediction performance with PedGNN.}
\label{tab:results-synthToreal}
\begin{center}
\footnotesize
\begin{tabular}{@{}ccccccc@{}}
\toprule
Train         & Test  & $N_F$ & Accuracy   & Precision & Recall      & F1-score \\ 
\midrule
JAAD          & JAAD  & 18    & \vb{80.32} & 84.72      & \vb{87.91} & 85.29 \\  
{\thedataset} & JAAD  & 32    & 78.59      & \vb{89.48} &  84.62     & \vb{85.45} \\ 
\midrule
PIE           & PIE   & 08    & \vb{68.11} & 66.98      & 68.36      & 69.81 \\
{\thedataset} & PIE   & 16    & 62.74      & \vb{69.37} & \vb{79.31} & \vb{74.01} \\ 
\bottomrule
\end{tabular}
\end{center}
\end{table}
\let\vb\undefined
\let\vs\undefined

\newcommand{\vb}[1]{\textbf{#1}} 
\newcommand{\vs}[1]{\textit{#1}} 
\begin{table}
\caption{Performance when combining different datasets for training PedGNN. J: JAAD, P: PIE, S: {\thedataset}.}
\label{tab:results-complementaries}
\begin{center}
\footnotesize
\begin{tabular}{@{}ccccccc@{}}
\toprule
Train     & Test  & $N_F$ & Accuracy   & Precision  & Recall      & F1-score \\ 
\midrule
J         & J     & 18    & 80.32      & 84.72      & 87.91       & 85.29 \\  
S         & J     & 32    & 78.59      & \vb{89.48} & 84.62       & 85.45 \\ 
J + P     & J     & 08    & 72.36      & 74.22      & 89.18       & 81.20 \\
J + S     & J     & 32    & \vb{86.22} & 77.35      & \vb{96.19}  & \vb{85.96} \\ 
J + S + P & J     & 08    & 74.41      & 76.73      & 88.00       & 81.98 \\
\midrule
P         & P     & 08    & 68.11      & 66.98      & 68.36       & 69.81 \\
S         & P     & 16    & 62.74      & 69.37      & 79.31       & 74.01 \\ 
P + J     & P     & 32    & 69.26      & 77.03      & 75.26       & 76.13 \\ 
P + S     & P     & 16    & \vb{70.52} & 74.80      & \vb{82.73}  & \vb{79.12} \\ 
P + S + J & P     & 08    & 69.34      & \vb{78.24} & 70.20       & 76.35\\ 
\bottomrule
\end{tabular}
\end{center}
\end{table}
\let\vb\undefined
\let\vs\undefined

\newcommand{\vb}[1]{\textbf{#1}} 
\newcommand{\vs}[1]{\textit{#1}} 
\begin{table}
\caption{{\thedataset} (S) as testing dataset. J: JAAD, P: PIE.}
\label{tab:synthfortesting}
\begin{center}
\footnotesize
\begin{tabular}{@{}ccccccc@{}}
\toprule
Train     & Test  & $N_F$ & Accuracy   & Precision  & Recall      & F1-score \\ 
\midrule
J         & S     & 08  & \vb{72.23}   & \vb{77.33} & 83.97       & 80.97 \\
P         & S     & 08  & 69.66        & 74.54      & 82.47       & 78.30 \\
J + P     & S     & 08  & 71.40        & 73.11      & \vb{85.74}  & \vb{82.98} \\
\midrule
J + S     & J     & 32  & 86.22        & 77.35      & 96.19       & 85.96 \\ 
P + S     & P     & 16  & 70.52        & 74.80      & 82.73       & 79.12 \\ 
\bottomrule
\end{tabular}
\end{center}
\end{table}
\let\vb\undefined
\let\vs\undefined

\begingroup
\newcommand{\vb}[1]{\textbf{#1}} 
\newcommand{\vs}[1]{\textit{#1}} 
\setlength{\tabcolsep}{2pt} 
\begin{table}
\caption{J: JAAD, P: PIE, S: {\thedataset}. GT refers to using ground truth skeletons from CARLA. PedGraph+$^\star$ refers to PedGraph+ \cite{cadena2022pedgraph+} but only considers pedestrian skeletons (from AlphaPose) as input information ($N_F=32$ is used in \cite{cadena2022pedgraph+}). For PIE, PedGraph+$^\star$ only considers the $\sim30\%$ of C/NC cases, which we denote as P*.}
\label{tab:results-pedgnnassesment}
\begin{center}
\footnotesize
\begin{tabular}{@{}lccccccc@{}}
\toprule
Model             & Train  & Test   & $N_F$ & Accuracy   & Precision  & Recall     & F1-score \\ 
\midrule
PedGNN            & S (GT) & S (GT) & 08    & 89.29      & 95.85      & 88.69      & 92.14 \\ 
\midrule
PedGNN            & J      & J      & 18    & 80.32      & \vb{84.72} & \vb{87.91} & \vb{85.29} \\  
PedGraph+$^\star$ & J      & J      & 32    & \vb{83.85} & 53.76      & 59.21      & 56.36 \\ 
\midrule
PedGNN            & P      & P      & 08    & 68.11      & 66.98      & \vb{68.36} & \vb{69.81} \\
PedGraph+$^\star$ & P*     & P*     & 32    & \vb{79.15} & \vb{77.91} & 36.51      & 49.72 \\ 
\bottomrule
\end{tabular}
\end{center}
\end{table}
\let\vb\undefined
\let\vs\undefined
\endgroup

\newcommand{\vb}[1]{\textbf{#1}} 
\newcommand{\vs}[1]{\textit{#1}} 
User
\begin{table}
\caption{Memory footprint and inference time of PedGNN and different models from the state-of-the-art. All times are computed on an NVIDIA GTX 1080 GPU. We have extracted these times from the respective papers.}
\label{tab:memoryspeed}
\begin{center}
\setlength{\tabcolsep}{7pt}
\begin{tabular}{@{} l c r @{}}
\toprule
            Model                       &            Size (MB)        & Inference time (ms)\\
\midrule
PCPA \cite{kotseruba2021benchmark}      &       118.8                 & 38.6 \\
Global PCPA \cite{yang2021predicting}   &       374.2                 & 70.83\\
FUSSI \cite{piccoli2020FuSSINetFO}      &       8.4                   & 34.92 \\
PedGraph \cite{cadena2019pedgraph}      &       0.22                  & 29.01\\
PedGraph+ \cite{cadena2022pedgraph+}    &       0.28                  & 5.47 \\
TEP \cite{achaji2022isattention}        &       12.8                  & 2.85 \\
V-PedCross \cite{bai2022DeepVD}         &       4.8                   & -\\
PedGNN (Ours)                           &       \textbf{0.027}        & \textbf{0.58}\\
\bottomrule
\end{tabular}
\end{center}
\end{table}
\let\vb\undefined
\let\vs\undefined
\let\vb\undefined
\let\vs\undefined

\subsection{Experiments and discussion}
We start the experiments by evaluating how effective {\thedataset} is training our PedGNN model to perform on the JAAD and PIE testing sets. Table \ref{tab:results-synthToreal} shows the results. The $N_F$ value refers to the number of input frames (per pedestrian skeletons) that was best for each case according to the previously described training protocol. Training on {\thedataset} requires considering more frames than using the respective JAAD/PIE training data. However, this does not affect the prediction latency since, as we have mentioned before, we use a temporal sliding window of a 1-frame step. Moreover, in an NVIDIA GTX 1080 GPU, for  $N_F=32$ PedGNN only takes $\sim0.6$ms to perform the inference. In terms of accuracy, training on the respective real-world datasets is better than training on {\thedataset}. However, we can see how it is not the case for F1-score, which is a more unbiased metric than accuracy when the testing data distribution presents a class unbalanced. Table \ref{tab:labelsdataset} shows that this is the case here since both JAAD and PIE have testing sets clearly biased toward the crossing (C) class. Thus, we think {\thedataset} is an effective dataset for training C/NC prediction models. 

At this point, it is worth commenting that, as shown in Table \ref{tab:statsdataset}, {\thedataset} provides a skeleton GT for each pedestrian (coming from the CARLA simulator). The joints used by PedGNN are a subset of those provided by the CARLA simulator, so the mapping is straightforward. Moreover, fitting confidences $c_j$ can be set to 1 since skeletons are perfectly fitting pedestrians. Therefore, this raises the question of using such skeletons as GT for training with {\thedataset} instead of applying AlphaPose to the synthetic pedestrians. We did the corresponding experiments, however, F1-score dropped $\sim10$ points when testing on JAAD and $\sim6$ for PIE. In other words, a synth-to-real domain gap is induced by the use of different skeleton sources at training (GT) and testing (Alphapose) time. We leave future work to investigate more in deep the underlying reasons for the domain gap and keep using AlphaPose for all the training runs involving {\thedataset}.

Sometimes we may have real-world training data labeled for C/NC prediction, as is the case of PIE and JAAD. Then, it is also interesting to see if the synthetic data at hand can act as a complement, giving rise to better-performing models. Note that by using the same skeleton fitting method we avoid the synth-to-real domain gap provided this method performs well in the real and synthetic domains. According to our experiments, AlphaPose fulfills so. Therefore, we have combined {\thedataset} training data with JAAD and/or PIE training data for assessing the complementarity of these datasets. Table \ref{tab:results-complementaries} presents the corresponding results, where we also include those in Table \ref{tab:results-synthToreal} for easier comparison. We can see that, when testing in JAAD, combining JAAD and {\thedataset} gives better results than using JAAD or {\thedataset} alone and even than combining JAAD and PIE. In fact, it seems that PIE produces negative transfer when combining the three datasets. On the other hand, since the testing set of PIE has around ${4\times}$ more C-frames than JAAD, and around ${9\times}$ more NC-frames, results on PIE are of special interest. We can see that, when testing in PIE, combining PIE and {\thedataset} gives rise to the best results in terms of accuracy and F1-score. Thus, we think that {\thedataset} can complement real-world datasets for training purposes.

\begin{figure}[t!]
    \begin{subfigure}[b]{\columnwidth}
    \centering
        \includegraphics[width=0.7\columnwidth]{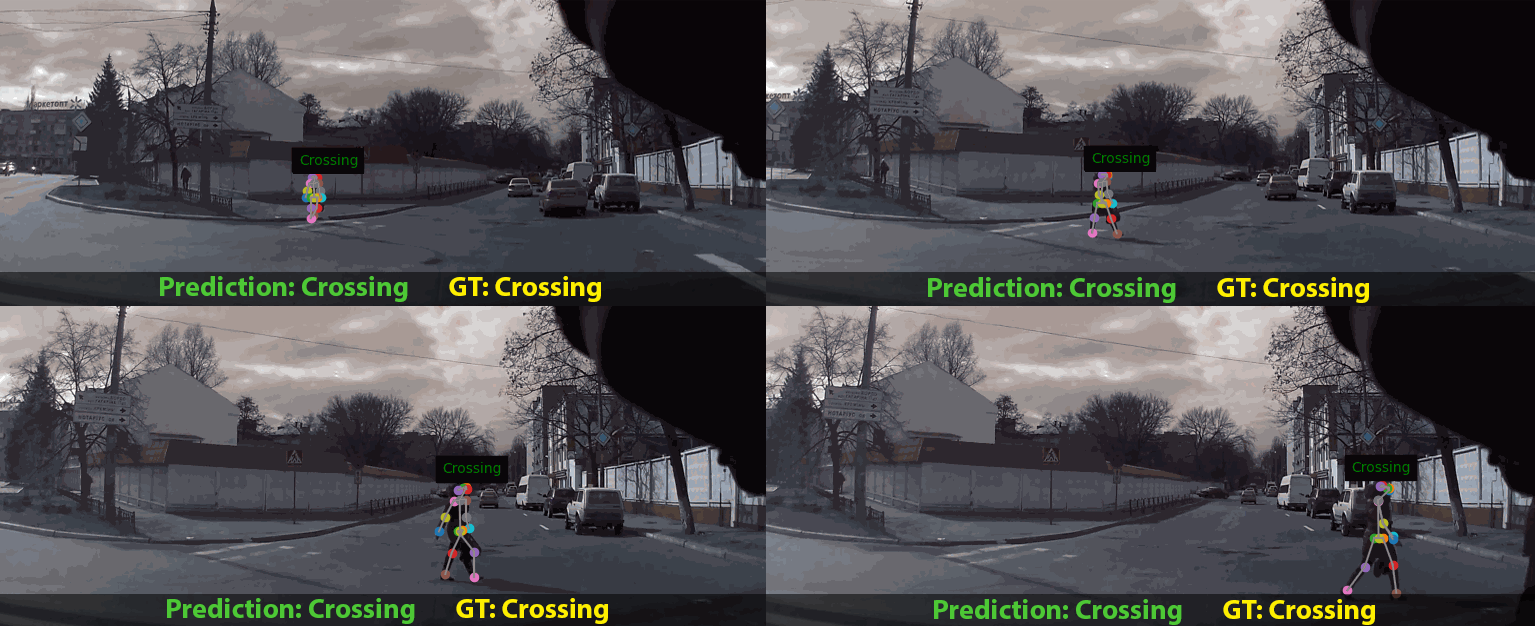}
        \caption{}
    \end{subfigure}
   
    \begin{subfigure}[b]{\columnwidth}
    \centering
        \includegraphics[width=0.7\columnwidth]{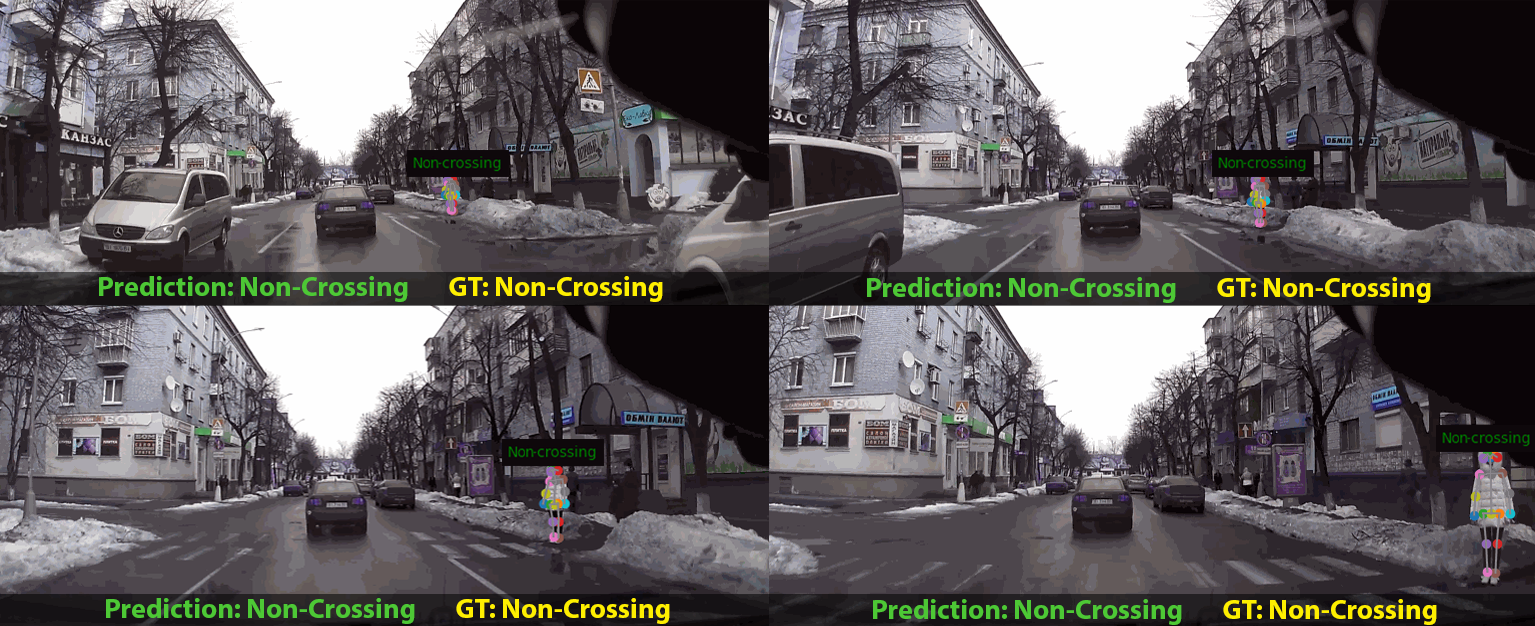}
        \caption{}
    \end{subfigure}
    
    \begin{subfigure}[b]{\columnwidth}
    \centering
        \includegraphics[width=0.7\columnwidth]{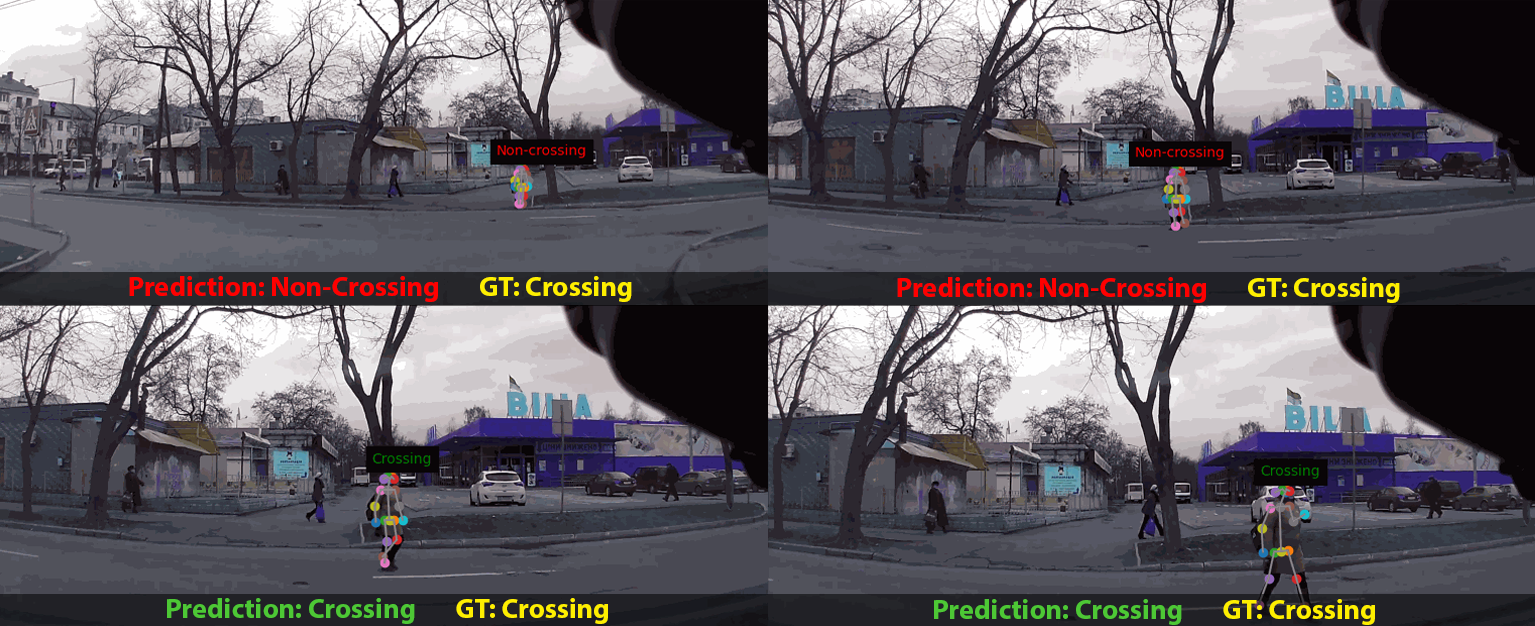}
        \caption{}
    \end{subfigure}
    
    \begin{subfigure}[b]{\columnwidth}
    \centering
        \includegraphics[width=0.7\columnwidth]{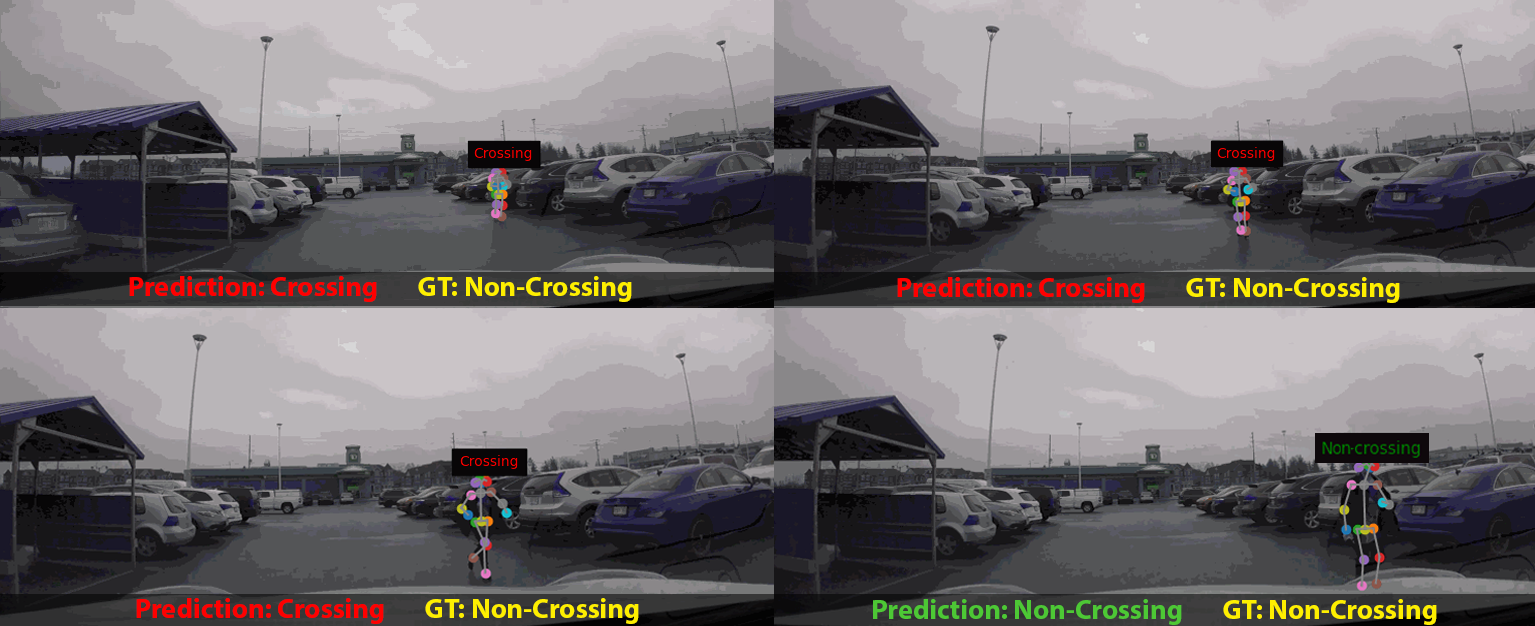}
        \caption{}
    \end{subfigure}
    
    \caption{Performance of PedGNN trained on JAAD+{\thedataset} and tested on JAAD. Cases (a) and (b) are fully successful, while in cases (c) and (d) there are C/NC prediction discrepancies with the labels provided by human labelers (GT). Time in each sequence runs from top-left to bottom-right.}
    \label{fig:results-qualitative}
\end{figure}

We can also assume that we use {\thedataset} for testing purposes. Table \ref{tab:synthfortesting} shows the results corresponding to training with the real-world training sets and testing in the {\thedataset} testing set. For easier visual comparison, we also include the best models obtained when testing in real-world testing sets. These correspond to training on the respective training set plus the training set of {\thedataset}. We can see that performance metrics report comparable values when testing in real-world sets and in our synthetic set. Thus, we think {\thedataset} can play the role of the testing set too; in other words, it is not easier than their real-world counterparts.

At this point, we put the focus on the PedGNN model. On the one hand, we assess its potential by using {\thedataset} and the associated pedestrian GT skeletons, so that results are not influenced by the skeleton fitting method in place (here AlphaPose). Moreover, we compare our results with the state-of-the-art method on C/NC prediction here named PedGraph+ \cite{cadena2022pedgraph+}. Table \ref{tab:results-pedgnnassesment} shows the results. For PedGraph+ we have copied the results reported in \cite{cadena2022pedgraph+} when only AlphaPose-based skeletons are considered as input. However, for the PIE dataset, only a portion of the data is considered in \cite{cadena2022pedgraph+}, roughly the $30\%$. We can observe that PedGNN has great potential of providing good performance, which can be seen when training and testing with perfect pedestrian skeletons (GT). Note that F1-score is $\sim92\%$. Of course, there is room for improvement. Compared to PedGraph+ assuming the same input data (AlphaPose-based skeletons), we can see how PedGraph+ performs better in terms of accuracy, but significantly worse when using the more representative F1-score metric. Moreover, we can see in Table \ref{tab:memoryspeed} how in terms of memory footprint and inference speed PedGNN is significantly more lightweight and faster. 

Finally, as an example of qualitative results, Fig. \ref{fig:results-qualitative} illustrates the performance of PedGNN trained on JAAD+{\thedataset} and tested on JAAD. In case (a), while the ego-vehicle turns to the right, 
the intention of a pedestrian that started to cross in the left 
is properly predicted from the very beginning. In case (b), while the ego-vehicle 
moves straight forward, a pedestrian standing still at the border of the road
is properly predicted as a non-crossing pedestrian. 
In cases (c) and (d), PedGNN requires more frames to reach the proper prediction. 
In case (c) the pedestrian seems to take the crossing decision later than in case (a), so predicting the intentions required some additional time. In case (d), the pedestrian seems to start crossing in front of the ego-vehicle in a parking area, but finally, it does not. In fact, for us it is unclear if the GT is right, after all, it is based on human labelers and, therefore, there is subjectivity. For instance, if the initial frames have been labeled after looking at what the pedestrian did at the final ones, this would be like using the future to predict the present/past, which cannot be done by the temporal sliding window mechanism used to process onboard continuous image sequences.

\section{Conclusion}
\label{sec:conclusion}


In this paper, we have introduced our framework {\thefw} which allows the generation of synthetic datasets labeled for the pedestrian C/NC prediction task. It works on top of the CARLA simulator so being aligned with the autonomous driving research community. Advanced users can programmatically design their pedestrian C/NC scenarios, the rest can adjust a configuration file to use existing scenarios. For example, we have generated the {\thedataset} dataset by using {\thefw}. It is diverse and contains a large amount of pedestrian C/NC cases. We have shown its usefulness by running an extensive set of experiments. We have seen that it can play the role of the training set alone, it can complement real-world training sets, and it can play the role of the testing set. Most experiments are based on our model PedGNN, also introduced in this paper. It processes sequences of pedestrian skeletons to produce C/NC predictions. Our experiments show that PedGNN produces state-of-the-art results, despite being significantly more lightweight and faster than previous C/NC prediction models. In future work, we plan to use {\thedataset} and PedGNN to address the synth-to-real unsupervised domain adaptation problem for the pedestrian C/NC prediction task.









\bibliographystyle{ieee_fullname}
\bibliography{bib/references}

\end{document}